\documentclass{article}
\usepackage[latin9]{luainputenc}
\usepackage{amsmath}
\usepackage{amssymb}
\usepackage{graphicx}
\usepackage{xcolor,comment}
\usepackage{enumitem}
\usepackage{hyperref}

\makeatletter
\usepackage{spconf}

\title{FULLY CONVOLUTIONAL FRACTIONAL SCALING}
\name{Michael Soloveitchik \&    Michael Werman\thanks{Thanks to DFG for funding.}}
\address{Computer Science\\The Hebrew University of Jerusalem}
\makeatother

\begin{document}
\topmargin=0mm 
\maketitle 
\begin{abstract}
We introduce a  fully convolutional fractional scaling component, FCFS.
Fully convolutional networks can be applied to any size input and  previously did not support non-integer scaling.  Our architecture is simple with an efficient single layer implementation. Examples and code implementations of three common scaling methods are published.
\end{abstract}

\section{Introduction}
\label{sec:intro}
Image scaling is a ubiquitous image processing operation. Neural networks 
based only on convolutions have the nice property that they can be
applied to any size object. This  family of architectures is
named FCN, Fully Convolutional Networks.  Many FCN models consist of up/down-sampling layers, albeit with integer factors. Here we present a fully convolutional
fractional scaling component for CNNs,  FCFS.

Various tasks, such as \textbf{instance\textbackslash semantic segmentation} \cite{dai2016instance} and \cite{long2015fully},
\textbf{style transfer} \cite{zhu2017unpaired}, \textbf{super-resolution} \cite{yamanaka2017fast}, \textbf{image compression} \cite{yagnasree2021image},
\textbf{satellite image segmentation} \cite{garnot2021panoptic}, \textbf{general object detection} \cite{wang2017video}, etc. have state-of-the-art solutions
based on integer scale up/down-sampling layers embedded in FCN architectures.

Up/down scaling architectures are extensive in computer vision fields. 
Zhang et al. \cite{zhang2020residual} used the well known Laplacian Pyramids, together with a deep neural network to train a super-resolution model.
Luo et al. \cite{Luo_2021_CVPR} the authors used an up-sampling sequence of layers to find the optical-flow of an image. 
Various super-resolution models as \cite{shocher2018zero}, \cite{pons2021upsampling} and \cite{parihar2022video} tried to find the proper HR-counterpart of an LR-image when it's acquisition isn't predetermined. Maeda et al. \cite{maeda2020unpaired}   implemented an unpaired super resolution model based on cycle-consistency and up/down scaling layers.
Saeedan et al.\cite{saeedan2018detail} succeeded to preserve important image details during down scaling with average-pooling layers. In addition \cite{li2018multi} reported success on using 
using the $pixelshuffle$ component to up-scale an input image by any integer scaling integer factor. 

Audio processing  also uses scaling methods,  \cite{yuan2018unsupervised} studied the appearance of artifacts on audio signals after applying up-sampling methods.

Works like \cite{schlegl2017unsupervised} and \cite{le2021brain2pix} used scaling layers and FCN to detect anomalies in medical images and brain 3D reconstruction respectively.

In this paper, we suggest a fully convolutional fractional, generalizing the integer, scaling component. Our architecture has an elegant and simple single layer implementation that allows easy integration in any FCN. Implementations of three common scaling methods: "nearest-neighbour", "bilinear" and "bicubic" interpolation can be found in  \href{https://github.com/Michael-Soloveitchik/FCFS}{Project Page}

\section{Previous Work}
\label{sec:format}

The fractional convolutional scaling was proposed, first time, in \cite{graham2014fractional}, and later have been  exceeded by \cite{Zhai_2017_CVPR}. It has stochastic implementations. Their works randomly pools the input with overlapping patches to achieve the desired fraction.  Afterwards \cite{hang2017bi} suggested bilinear average pooling, their work is applicable only for scaling factors (denoted by f) in range $1\leq f\leq 2$. All the former mentioned work aimed only for down-scaling tasks.

Another work is
 \cite{chen2021convolutional}, their motivation was compressing video and they 
suggested an architecture with fixed input and output shapes. 

Our solution is not restricted to fixed sizes and doesn't
use stochastic sampling, providing a clean fully convolutional component to perform fractional down/up scaling.

\section{Approach}

For simplicity, we develop the theory for 1D tensor convolutions.
Then we  present the generalization  for the higher dimensional cases.

{\bf 1D discrete tensor convolution:} Given an array $x$  and  a convolution kernel $h$ of size $2K+1$,  the convolution of $x$ and $h$ is: 
\[
(x*h)[i]=\sum_{k= -K}^{K}x[i-k]\cdot h[k]
\]

\subsubsection{Stride, Padding \& Pixelshuffle}

Our algorithm is based  on  three  operations:
\textbf{stride}, \textbf{padding}, and \textbf{pixelshuffle}.

{\bf Padding:}
Given an array $x$. Padding by $2p$ is the concatenation
$c^p |x | c^p$.
There are  other padding methods including reflection, zeros, and repetition of edge pixels.

\begin{figure}[h]
    \centering
\includegraphics[width=5cm]{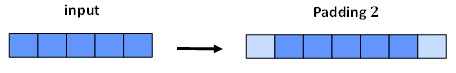}  \caption{Padding an  array of length $5$ by  $2$.}
  \label{fig:for}%
\end{figure}

{\bf Stride:} Given an array $x$ and a  convolution kernel $h$, the convolution of $x$ and $h$ with stride $s$ is
\[
(x*_{s}h)[i]=(x*h)[s\cdot i]
\]
\begin{figure}[h]
    \centering
{\includegraphics[width=5cm]{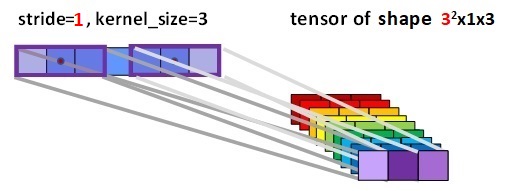}}
\caption{1D Tensor convolution with padding=3, stride=$4$ out\_channels=$3$ on a length 5  array.}
\end{figure}

{\bf Pixel Shuffling:}
Let the pixelshuffle be $r$, given a tensor of shape $r \times N$ pixelshuffle \cite{shi2016real} reshapes and rearranges  it's elements in a tensor of shape $rN$.
\begin{figure}[h]%
\centering{\includegraphics[width=6.0cm]{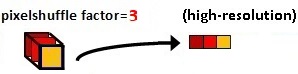}}
\caption{Pixelshuffle of $3 \times 1$ tensor returning a tensor of shape 3.}
\end{figure}
\subsection{Fully Convolutional Fractional Scaling}
The architecture we purpose, FCFS, carries out fractional scaling. 
{\bf \textbf{F}ully \textbf{C}onvolutional
\textbf{F}ractional \textbf{S}caling}:
Given a real array $x$ and a scaling factor $\frac{r}{s}$, we define the following algorithm.\\

\label{1D_FCFS_pseudocode}
\textbf{ FCFS(input}: Tensor) $\rightarrow$ Tensor:

\ \ \ \ \ x = pad(x, padding=$2K$)

\ \ \ \ \ x = conv(x, out\_channels=$r$, stride=s, 

\ \ \ \ \ \ \ \ \ \ \ \ \ \ \ \ \ \ \ \ \ \ \ \ 
kernel\_shape=2K+1, kernel\_weights=W)

\ \ \ \ \ {\bf return} pixelshuffle(x, factors=$r$)

\subsubsection{Description}
\label{subsub:Description}
The scaling is relative to the padded tensor of shape $N$.
The architecture contains only a single hidden layer, whose shape is $r \times \frac{N}{s}$. Given a scale factor $\frac{r}{s}$ we apply a convolution with $stride=s$. Each of the $r$ convolution kernels produces an interpolation for offset  $i+\frac{j-1}{r}\ \ s.t.\ j\in[1,..,r]$.  Thus the hidden layer  is of shape $r\times \frac{N}{s}$. Applying $pixelshuffle$  results in  an array with the desired shape of $\frac{r}{s} N$

The parameters $K$, $kernel\_shape$, and $kernel\_weights$ depend on the interpolation method see  examples  section \ref{sec::exambles}. The hidden layer's shape is bilinearly dependent on  output shape and $r$, which is the space and time complexity of the component. 

\section{2D \& ND Extensions}
The adaptations needed for 2D and ND are straightforward. Special attention needs  to be paid to  $Pixelshuffle$.

\subsection{ND Pixelshuffle}
\begin{figure}[h]
\centering{\includegraphics[width=5cm]{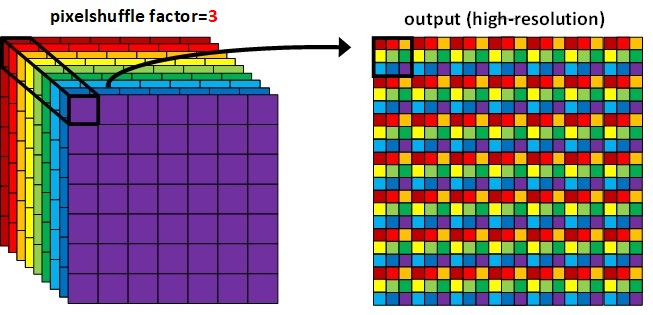}}
\caption{Pixelshuffle of a $3^{2}\times7\times 7$ tensor to a $3\cdot7\times 3\cdot7$ matrix.}
\label{scheme}
\end{figure}

Here we propose a slight generalization of Pixelshuffle.\\
{\bf PixelShuffle:}
For $n \ge 2$ and a tensor of shape
$(r_1\cdot...\cdot  r_i\cdot...\cdot r_n)\times N_1\times...\times N_i\times...\times N_n$, Pixelshuffle rearranges the elements to a new tensor of shape $r_1N_1\times ...\times r_iN_i\times...\times r_nN_n$
    \[
        Out[i_1,..., i_i, ..., i_n] = x[r
        , \lfloor\frac{i_1}{r_1}\rfloor,..., \lfloor\frac{i_i}{r_i}\rfloor,... , \lfloor\frac{i_n}{r_n}\rfloor]\]
        \[
        r = \sum_{t=0}^{n-1}( \prod_{j=1}^{n-t-1}r_j)((i_{n-t}-1) ~{mod}~\ r_{n-t})
    \]
Figure \ref{scheme}  illustrates  the  formula.

\subsection{ND Fully Convolutional Fractional Scaling}

\label{ND_FCFS_pseudocode}

To scale an $ND$ input signal by  scaling factors: $\frac{r_{1}}{s_{1}},...,\frac{r_{n}}{s_{n}}$  for the different  dimensions.

\textbf{ FCFS(input}: Tensor) $\rightarrow$ Tensor:

\ \ \ \ \ x = pad(x, padding=$2K_1,\cdots,2K_n)$

\ \ \ \ \ x = conv2d(x, out\_channels=$\prod_{i} r_i$, 

\ \ \ \ \ \ \ \ \ \ \ \ \ \ \ \ \ \ \ \ \ \ \ \ stride=[$s_{1},...,s_{n}$], 

\ \ \ \ \ \ \ \ \ \ \ \ \ \ \ \ \ \ \ \ \ \ \ \ 
kernel\_shape=[$2K_{1}+1,...,2K_{n}+1$],

\ \ \ \ \ \ \ \ \ \ \ \ \ \ \ \ \ \ \ \ \ \ \ \ kernel\_weights=W

\ \ \ \ \ {\bf return} pixelshuffle(x, factors=[$r_{1},...,r_{n}$])



\section{EXAMPLES}
\label{sec::exambles}
\subsection{Illustration of FCFS}

\begin{figure}[h]
\centering{\includegraphics[width=8.5cm]{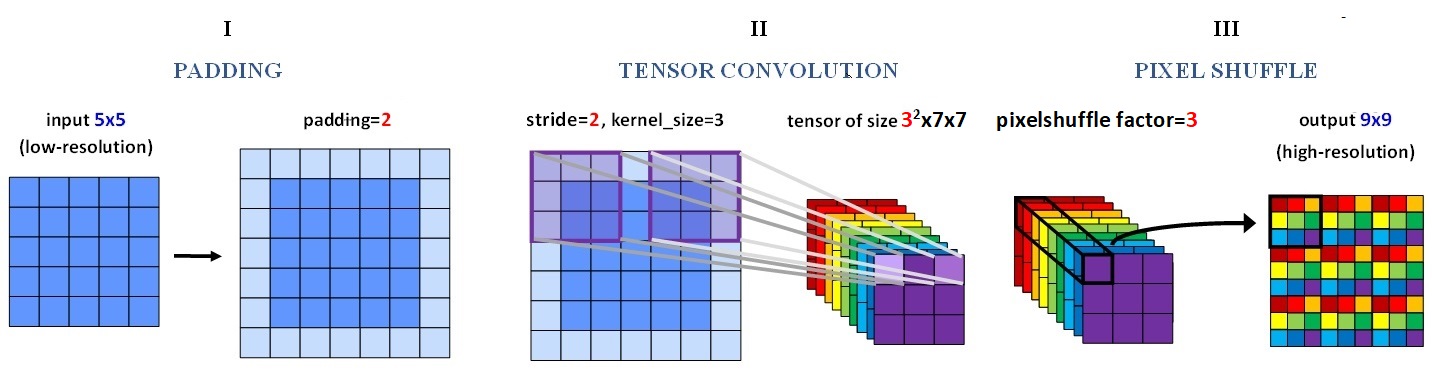}}
\caption{Illustration of 2D  $\frac{3}{2}$ \textbf{F}ully \textbf{C}onvolutional \textbf{F}ractional \textbf{S}caling.}
\label{FCFS illustration}
\end{figure}

Figure \ref{FCFS illustration} illustrates  FCFS  on a $5\times5$ image  with a $\frac{3}{2}$ up-sacle factor.
The output image is     a $9\times9$ image, 
\[
9\times9=\frac{3}{2}\cdot(5+1)\times(5+1)\Longrightarrow\ output=\frac{3}{2}\cdot input
\] as expected.

\subsection{Convolution's kernel weights}
$FCFS$ supports various scaling-methods through  the parameters.  In this section, we present   kernel weights for various image scaling-methods.\\

Consider  $f=\frac{3}{2}=\frac{r}{s}$ scaling.
According to the offsets  described in \ref{subsub:Description}, we have $3^2=9 (=r^2)$ different kernels. We present the kernel of offsets $(1,1)$ and $(1,3)$ denoted by $W_{1,1}$ and $W_{1,3}$.

\subsubsection{Nearest neighbour interpolation}
From \cite{trusov2020analysis} nearest neighbour:
\[
W_{1,1}:=\left[\begin{array}{cc}
1. & 0.0\\
0.0 & 0.0
\end{array}\right]~~W_{1,3}:=\left[\begin{array}{cc}
0.0 & 1.0\\
0.0 & 0.0
\end{array}\right]
\]

\subsubsection{Bilinear interpolation}
 From \cite{trusov2020analysis}, bilinear interpolation which is based on the 4 nearest pixels around the point of interpolation:
\[
W_{1,1}:=\left[\begin{array}{cc}
0.44 & 0.22\\
0.22 & 0.11
\end{array}\right]~~W_{1,3}:=\left[\begin{array}{cc}
0.22 & 0.44\\
0.11 & 0.22
\end{array}\right]
\]

\subsubsection{Biqubic interpolation}
 From \cite{trusov2020analysis} bicubic interpolation as derived from the formula, published in \cite{trusov2020analysis}:
 \small
 \[
 {\displaystyle W(\Delta)={\begin{cases}1.5|\Delta|^{3}-2.5|\Delta|^{2}+1&{\text{for }}|\Delta|\leq 1,\\-0.5|\Delta|^{3}+2.5|\Delta|^{2}-4|\Delta|-4a&{\text{for }}1<|\Delta|<2,\\0&{\text{otherwise}},\end{cases}}}
 \]
 \normalsize

 $\Delta=x-i,y-j$. Where the $x,y$ the subpixel point of interpolation and $i,j$ are the integer coordinates of the input image.
 \small
\[
W_{1,1}:=\left[\begin{array}{ccc}
0.16 & 0.16 & 0.07\\
0.16 & 0.16 & 0.07\\
0.07 & 0.07 & 0.03
\end{array}\right]~~
W_{1,3}:=\left[\begin{array}{ccc}
0.13 & 0.13 & 0.13\\
0.13 & 0.13 & 0.13\\
0.05 & 0.05 & 0.05
\end{array}\right]
\]
\normalsize

\section{Experiments}
To test the time complexity and quality of FCFS we ran the following experiments:
\begin{enumerate}[noitemsep,topsep=0pt,parsep=0pt,partopsep=0pt]
    \item We 
    compared running times of FCFS to  $torch.resize$ \cite{collobert2002torch}  for various scaling factors.
    \item We computed two commonly used metrics, PSNR \cite{hore2010image} and SSIM \cite{hore2010image}. again comparing FCFS  to  $torch.resize$ \cite{collobert2002torch}, for various scaling factors.  
\end{enumerate}

\subsection{Empirical methods}
\subsubsection{Scaling methods}
Each experiment was repeated 100 times. We tested for the three different  $scaling-methods$:
"nearest neighbours", "bilinear-interpolation" and "bicubic-interpolation" The weights were implemented as described in section \ref{sec::exambles}.
For each method, six up-scaling factors and six down-scaling factors were tested

\subsubsection{Hardware \& Datasets}
The experiments were carried out on NVIDIA RTX2070 GPU. 

The dataset used was Celeb\_A \cite{liu2018large} with more than 200K celebrity
images with 1024x1024 pixel resolution.

\subsection{Results}
\subsubsection{Experiment results}
\begin{figure}[h]
\centering{\includegraphics[width=8.7
cm]{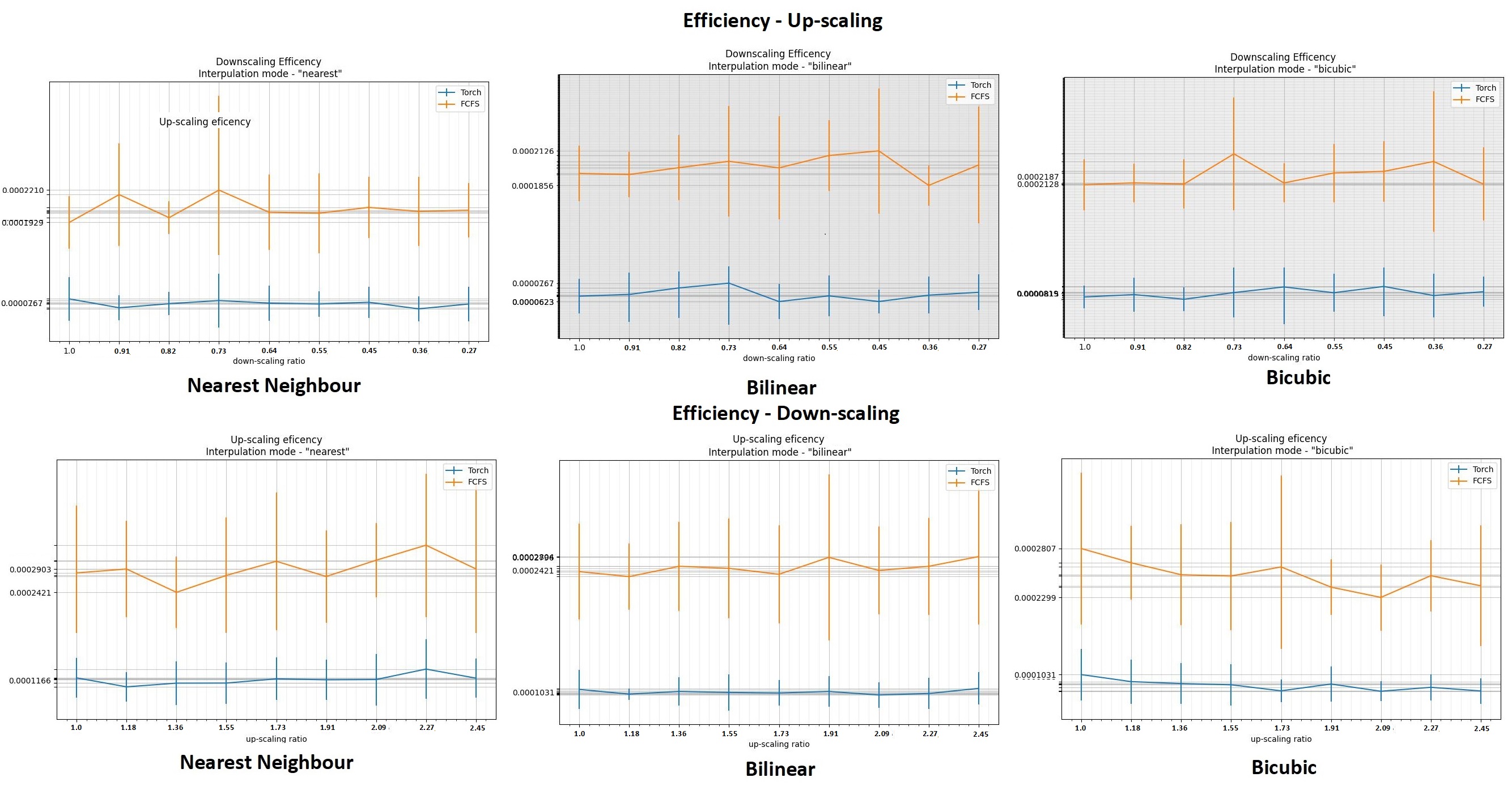}}
\caption{Efficiency graphs for up/down-scaling tasks $\times$ different $scaling-methods$}
\label{efficiency-graph.}
\end{figure}

The first experiment's running time  results are presented in figure \ref{efficiency-graph.}. No significant difference was found between  $torch.resize$ and  $FCFS$, neither in up-scaling nor in down-scaling. The $FCFS$ was 0.0003 seconds slower on average.

\begin{figure}[h]
\centering{\includegraphics[width=8.9
cm]{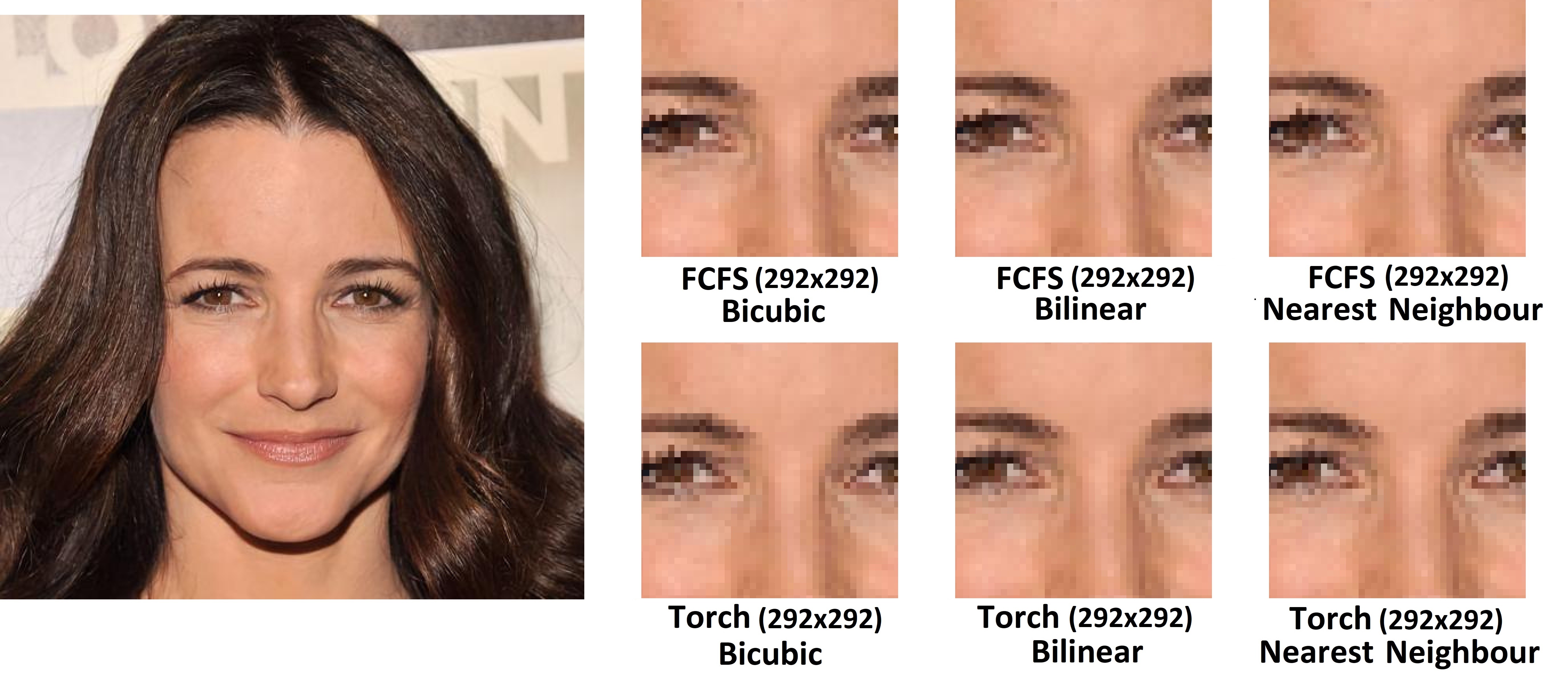}}
\caption{Down-scaling by factor =$\frac{2}{11}$}
\label{down-scaling-matrix}
\end{figure}

\begin{figure}[h]
\centering{\includegraphics[width=8.9 cm]{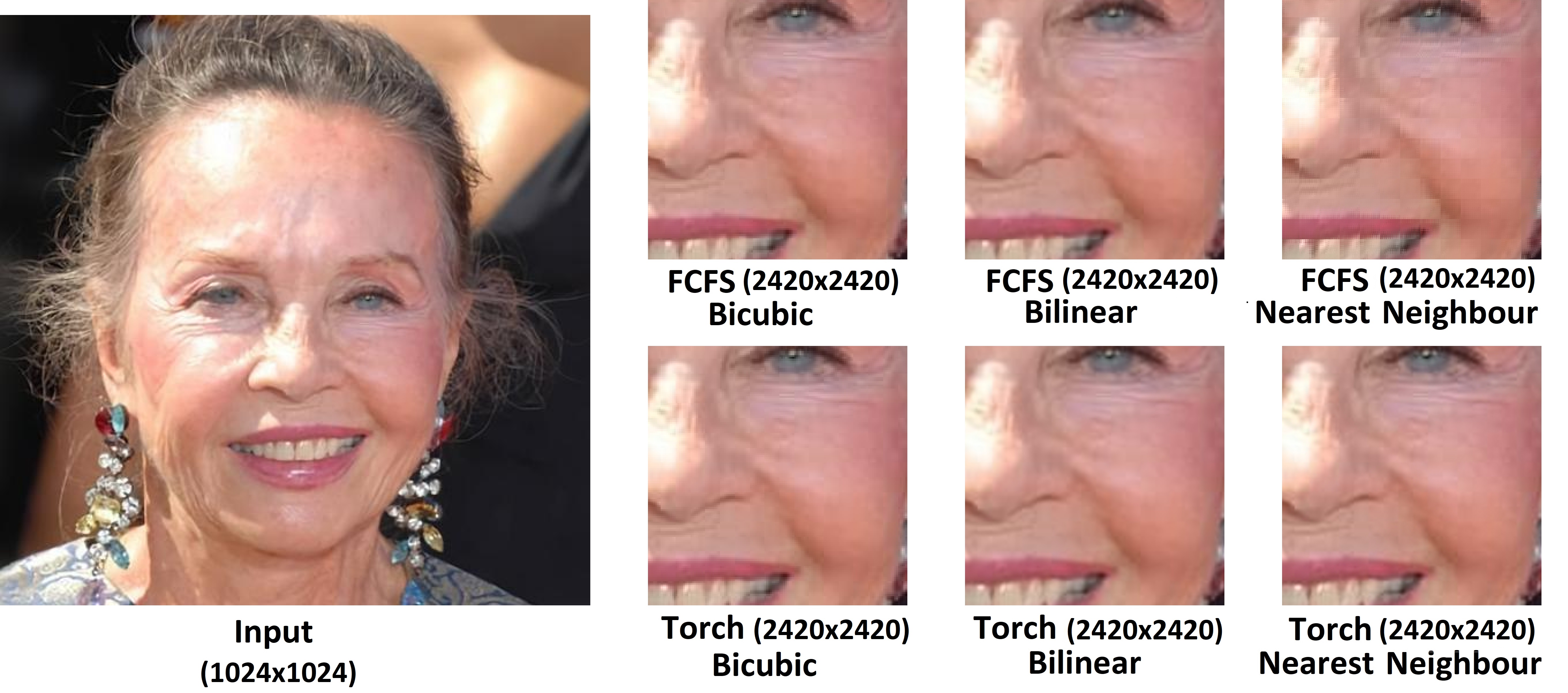}}
\caption{Up-scaling by factor =$\frac{27}{11}$.}
\label{up-scaling-matrix}
\end{figure}

The second experiment's visual sameness results are presented in figures \ref{down-scaling-matrix} and \ref{up-scaling-matrix}. Zooming in shows the visual artifacts that slightly differ between $FCFS$ and $torch.resize$.


\begin{figure}[h]
\centering{\includegraphics[width=8.9
cm]{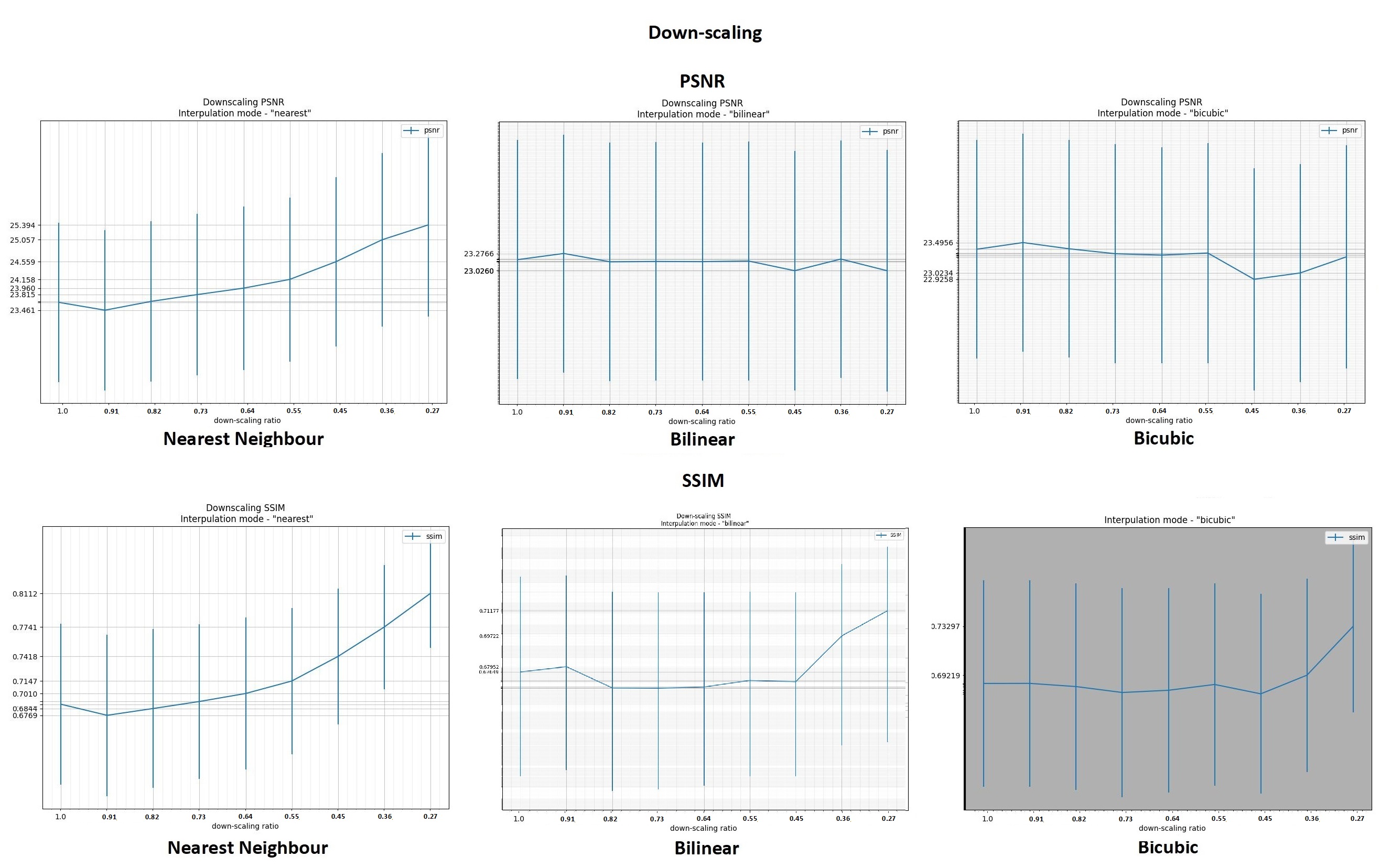}}
\caption{PSNR \& SSIM graphs for down-scaling tasks $\times$ different $scaling-methods$.}
\label{down-scaling-PSNR-SSIM}
\end{figure}

\begin{figure}[h]
\centering{\includegraphics[width=8.9
cm]{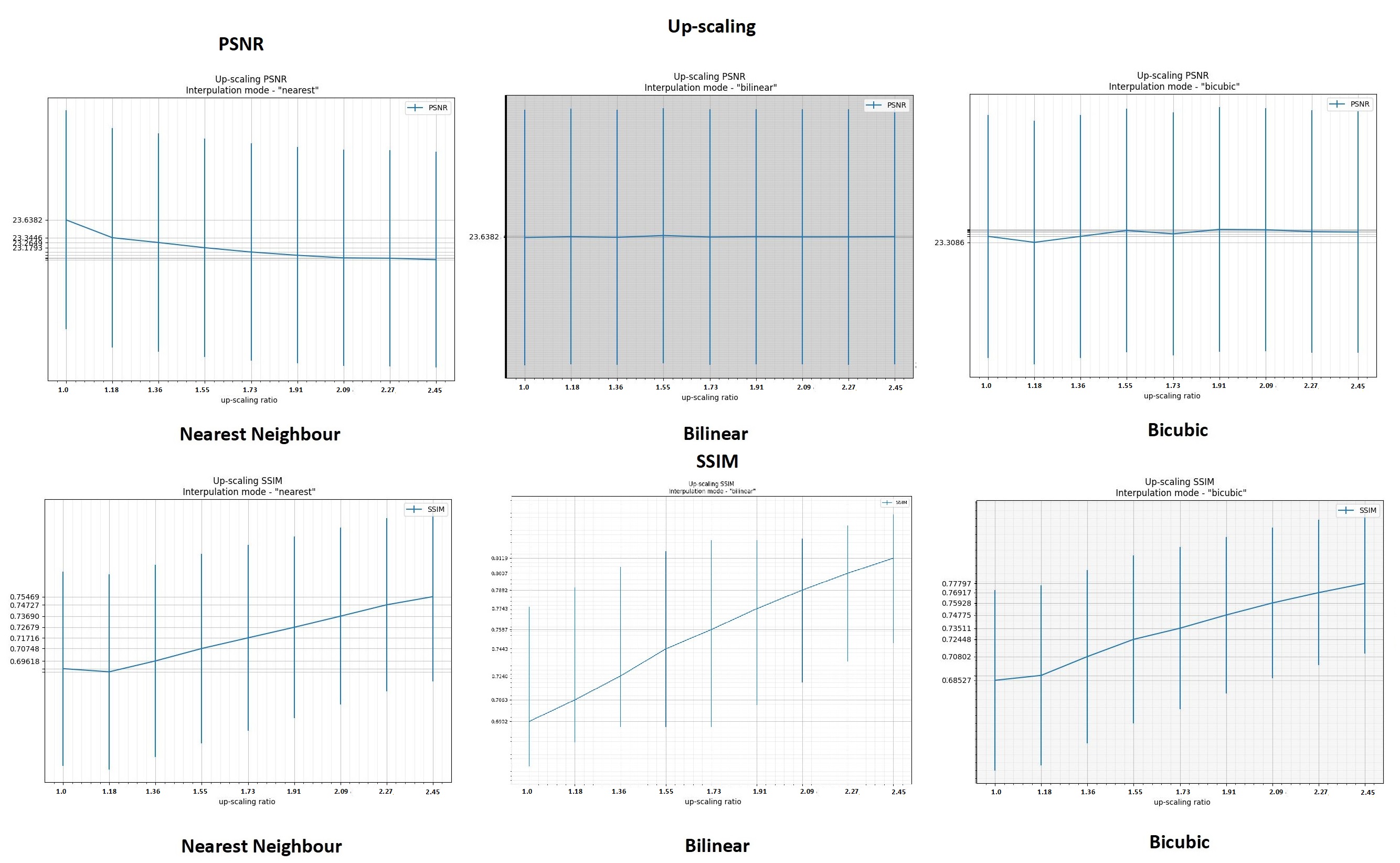}}
\caption{PSNR \& SSIM graphs for up-scaling tasks $\times$ different $scaling-methods$.}
\label{up-scaling-PSNR-SSIM}
\end{figure}

Figures \ref{down-scaling-PSNR-SSIM} and \ref{up-scaling-PSNR-SSIM} show distances between the output images for different $scaling-methods$.  PSNR and SSIM values above $20$ and close to $1.00$ respectively
support  visual sameness \cite{hore2010image}.
The  experiment shows  the  consistency of $FCFS$ with the $torch.resize$ implementation.   

\small

\section{SUMMARY \& FUTURE WORK}
\label{sec:summary}
We introduced  a fully convolutional fractional scaling component-$FCFS$ that is as efficient as the fixed shape scaling component ($torch.resize$). 

The benefit from a convolution based approach is the ability to learn  weights. FCFS allow to train the $kerenel\_weights$ and adjust them both in \textbf{shape} and \textbf{values} to the particular task. We aim to invest more effort in this direction in future work.

\clearpage 
\bibliographystyle{IEEEbib}
\bibliography{FCFS}

\end{document}